\documentclass[11pt,letterpaper]{article}
\usepackage[margin=1in]{geometry}
\usepackage[T1]{fontenc}
\usepackage{mathptmx}
\usepackage[numbers,sort&compress]{natbib}
\usepackage{float,placeins}
\usepackage[hidelinks]{hyperref}
\hypersetup{pdftitle={Consensus-Aware Multi-Source Fusion for Reference-Guided Camouflaged Object Detection},pdfauthor={Junyang Xia, Luocheng Zhang, Wenwen Pan, Chifeng Zhu, Yang Yang, Xinchun Liu, Jiajun Ding}}
\usepackage{amsmath,amssymb,booktabs}
\usepackage{graphicx}
\usepackage{xcolor}

\ExplSyntaxOn
\str_map_inline:nn {abcdefghijklmnopqrstuvwxyzABCDEFGHIJKLMNOPQRSTUVWXYZ}
  { \mathcode`#1 = \int_eval:n { `#1 } }
\ExplSyntaxOff
\DeclareSymbolFont{uprightgreek}{U}{eur}{m}{n}
\DeclareMathSymbol{\alpha}{\mathord}{uprightgreek}{"0B}
\DeclareMathSymbol{\beta}{\mathord}{uprightgreek}{"0C}
\DeclareMathSymbol{\eta}{\mathord}{uprightgreek}{"11}
\DeclareMathSymbol{\theta}{\mathord}{uprightgreek}{"12}
\DeclareMathSymbol{\lambda}{\mathord}{uprightgreek}{"15}
\DeclareMathSymbol{\xi}{\mathord}{uprightgreek}{"18}
\DeclareMathSymbol{\sigma}{\mathord}{uprightgreek}{"1B}
\DeclareMathSymbol{\phi}{\mathord}{uprightgreek}{"1E}
\DeclareMathSymbol{\epsilon}{\mathord}{uprightgreek}{"22}
\DeclareMathSymbol{\tau}{\mathord}{uprightgreek}{"1C}
\graphicspath{{figures/}}

\title{Consensus-Aware Multi-Source Fusion for Reference-Guided Camouflaged Object Detection}
\author{Junyang Xia$^{\dagger}$, Luocheng Zhang$^{\dagger}$, Wenwen Pan$^{*}$,\\
Chifeng Zhu, Yang Yang, Xinchun Liu, Jiajun Ding\\[5pt]
\small School of Computer Science, Hangzhou Dianzi University,\\
\small Hangzhou 310018, China\\[4pt]
\small $\dagger$ These authors contributed equally to this work.\\
\small $*$ Corresponding author: \texttt{wenwenpan@hdu.edu.cn}}
\date{}
\begin{document}
\raggedbottom
\maketitle
\begin{abstract}
Reference-guided camouflaged object detection aims to segment a target whose visual appearance closely resembles its surroundings by exploiting auxiliary reference samples. The task remains difficult because reference samples contain inconsistent target cues, while generic visual representations are not inherently aligned with the target specified by the references. To handle these problems, we present a consensus-aware multi-source fusion framework. Reference-Conditioned Dual-Backbone Fusion (RCDF) couples trainable PVTv2 query features with frozen DINOv3 representations and uses reference-conditioned correlation to select foundation-model evidence before multi-scale fusion. The framework also aggregates multiple references through cross-reference consensus aggregation and injects reference information at semantic depths matched to the query features. Extensive experiments demonstrate the effectiveness of the proposed method. The results further show that reference consensus, target-conditioned foundation features, and hierarchical decoding provide complementary improvements under the evaluation protocol. The source code will be made publicly available upon acceptance.
\end{abstract}
\noindent\textbf{Keywords:} Camouflaged object detection; reference-guided segmentation; feature fusion; visual foundation models

\section{Introduction}
\label{sec:introduction}

Camouflaged object detection (COD) aims to identify and segment objects whose appearance blends into the surrounding scene~\cite{fan2022concealed}. It supports species discovery and concealed-scene understanding~\cite{han2018survey,fan2023advances}, microscopic medical analysis~\cite{li2022trichomonas,li2021mvdi25k}, agricultural pest inspection in complex backgrounds~\cite{cheng2017pest}, industrial surface-defect inspection~\cite{luo2024cddnet}, and visual search in safety-critical settings~\cite{fan2023advances}. Conventional COD discovers plausible concealed foreground regions from a query image, whereas many practical searches focus on a specific target category, such as a particular species, pest, or defect class. Reference-guided camouflaged object detection (Ref-COD) addresses this need by using a small set of target examples to specify what should be segmented~\cite{zhang2025refcod}. Existing Ref-COD methods typically encode the reference set into a shared target representation and transfer it to the query branch through feature aggregation, semantic guidance, or cross-level alignment~\cite{zhang2025refcod,cheng2023mlkg,wu2025uat}. Despite this progress, Ref-COD still faces three main challenges. First, when a query contains multiple camouflaged objects or visually similar distractors, the model must identify and segment only the target category specified by the references; generic visual features alone may activate on semantically unrelated regions~\cite{zhang2025refcod,cheng2023mlkg}. Second, weak contrast, small target regions, thin structures, and ambiguous boundaries make fine-grained detail recovery difficult, particularly when reference information is injected without regard to the semantic depth of query features~\cite{sun2022bgnet,ji2023dgnet,wu2025uat}. Third, reference samples can differ substantially in pose, scale, appearance, and quality, so treating them as equally reliable may propagate inconsistent or misleading target cues~\cite{zhang2025refcod,wu2025uat}.

Existing methods have addressed these challenges to different extents. First, for target-specific localization, R2CNet derives a common representation from visual references and generates a pixel-level prior mask through dense reference--query comparison~\cite{zhang2025refcod}; UAT aligns visual reference and query features through cross-attention~\cite{wu2025uat}; and MLKG and CGCOD strengthen target semantics using multi-level language knowledge and class prompts, respectively~\cite{cheng2023mlkg,zhang2025cgcod}. These strategies improve semantic guidance, but they either summarize visual references into a common representation or rely on textual and class-level cues, without using visual references to select target-relevant evidence from frozen foundation features. Second, for detail recovery, R2CNet enriches multi-scale query features under the guidance of its referring mask~\cite{zhang2025refcod}, while UAT combines adjacent-layer semantics with a cross-attention encoder and models predictive uncertainty in its decoder~\cite{wu2025uat}. However, neither method explicitly matches spatial and vector reference priors to query stages according to semantic granularity or maintains reference-aware guidance throughout progressive decoding. Third, for reference reliability, R2CNet and UAT aggregate information from multiple visual references, yet neither explicitly estimates sample-wise reliability at multiple semantic levels~\cite{zhang2025refcod,wu2025uat}. Thus, target-conditioned foundation-feature selection, depth-matched detail reconstruction, and reliability-aware multi-reference aggregation remain unresolved within a unified framework.

\begin{figure}[!t]
    \centering
    \includegraphics[width=\textwidth]{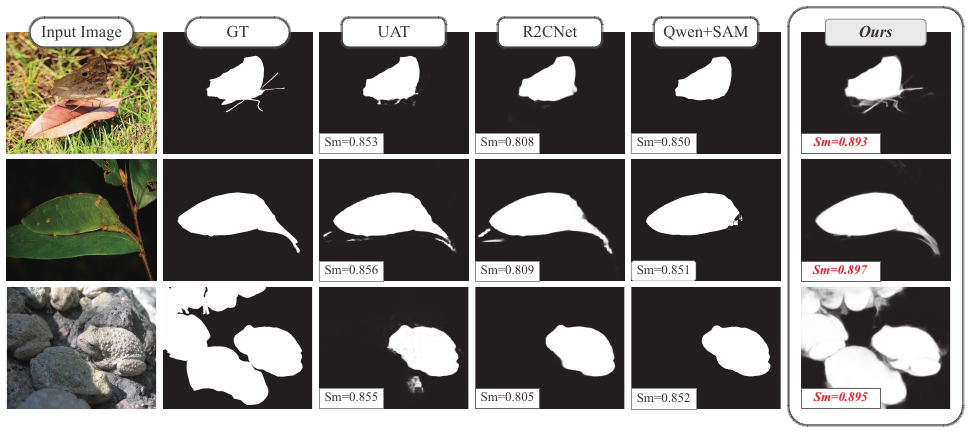}
    \caption{Representative results for reference-guided camouflaged object detection. Columns show the input, ground truth (GT), predictions from UAT, R2CNet, Qwen+SAM, and our method. The displayed $S_m$ values accompany examples with low contrast, thin structures, and multiple targets.}
    \label{fig:teaser}
\end{figure}

To address these limitations, we propose a consensus-aware multi-source fusion framework, which contains three complementary components: Reference-Conditioned Dual-Backbone Fusion (RCDF), hierarchical reference fusion and decoding (HRFD), and cross-reference consensus aggregation (CRCA). RCDF addresses target ambiguity in multi-object or distractor-rich scenes by forming a target descriptor from foreground-aware reference vectors and using it to select relevant frozen DINOv3 features before fusion with trainable PVTv2 query features. HRFD improves fine-grained detail recovery by injecting spatial and vector priors at matched semantic depths and progressively reconstructing target regions under region, auxiliary, and edge supervision. CRCA suppresses unreliable reference cues by learning reliability-weighted deep spatial, middle spatial, and global-vector priors from heterogeneous references. Figure~\ref{fig:pipeline} presents the complete architecture, while Figure~\ref{fig:teaser} illustrates the resulting advantages in representative scenes. In particular, the multi-object example shows that our method recovers multiple target instances more completely while suppressing unrelated regions, highlighting its ability to preserve target-specific localization in complex scenes.

The contributions of this work are summarized as follows.
\begin{itemize}
    \item We introduce RCDF, which couples trainable PVTv2 multi-scale features with frozen DINOv3 representations and uses reference-conditioned correlation to select target-relevant foundation-model evidence.
    \item CRCA estimates reference reliability to construct spatial and vector consensus priors from heterogeneous reference samples.
    \item We match spatial cross-attention and vector modulation to different query depths and progressively decode the conditioned features; a full-factorial evaluation verifies the individual and complementary effects of CRCA, RCDF, and HRFD.
\end{itemize}

\section{Related Work}
\label{sec:related-work}

\subsection{Conventional camouflaged object detection}

Conventional COD methods infer concealed regions from a single query image and differ mainly in how they expose weak foreground evidence. Bio-inspired coarse-to-fine search and iterative refinement progressively localize difficult targets~\cite{mei2021sinetv2,jia2022smr,zhang2022preynet}, while multi-scale and cross-level architectures exchange local detail with high-level context~\cite{pang2022zoom,chen2022context,pang2024zoomnext}. Boundary- and gradient-oriented models make structural discontinuities explicit through edge guidance, separated attention, feature decomposition, or gradient supervision~\cite{sun2022bgnet,zhu2022bsanet,he2023feature,ji2023dgnet}. Relation and ranking objectives further organize foreground--background interactions, object difficulty, and hierarchical tokens~\cite{zhai2021mgl,lv2021rank,yao2024hgit}. These families improve localization and contour recovery, but their target definition still comes entirely from the query image.

Other methods expand the evidence available to a query-only model. Frequency-aware networks separate or entangle spectral and spatial cues~\cite{cong2023frequency,sun2024fsel}; auxiliary depth can reveal object pop-out and support foundation-model adaptation~\cite{wu2023sourcefree,yu2024samdepth}; collaborative learning exploits related images~\cite{zhang2024collaborative}; and probabilistic or generative formulations model ambiguous regions and mask distributions~\cite{yang2021ugtr,chen2023diffusion}. Transformer decoders also strengthen long-range reasoning with shrinkage pyramids or masked separable attention~\cite{huang2023fspnet,yin2022camoformer}. Surveys and diagnostic studies summarize this progression from local appearance modeling toward richer structure and auxiliary cues~\cite{fan2022concealed,lv2023deeper}. Nevertheless, these cues generally help identify camouflaged foreground as a whole; they do not specify which concealed category a user intends to retrieve.

\subsection{Reference and auxiliary guidance}

Guided COD introduces information outside the query to resolve target ambiguity. Collaborative COD learns shared cues among semantically related images~\cite{zhang2024collaborative}, whereas class-guided COD uses category knowledge as a semantic constraint~\cite{zhang2025cgcod}. Ref-COD provides a more direct visual specification. R2CNet aggregates common representations from a set of salient reference objects and transfers them to a query branch~\cite{zhang2025refcod}. MLKG supplements alignment with progressively organized language knowledge from a multimodal large model~\cite{cheng2023mlkg}, while UAT combines referring-feature aggregation, cross-level attention, and probabilistic token modeling~\cite{wu2025uat}. These approaches demonstrate the value of target guidance, yet their auxiliary evidence is primarily summarized as a common or globally aligned signal. Reference sets can contain complementary local details as well as distractors; preserving this distinction requires both sample-wise reliability estimation and spatially resolved consensus.

More broadly, guided COD systems combine query hierarchies, visual references, external semantic features, and structural cues through concatenation, attention, or global aggregation. However, a single fusion operation does not resolve two distinct questions: which reference samples are trustworthy, and at what semantic depth should their information enter the query stream? Existing methods do not jointly model sample-wise reference reliability, spatially resolved consensus, and depth-matched reference injection, leaving these complementary aspects insufficiently explored.

\subsection{Transformers and visual foundation features}

The Transformer replaces recurrence with attention-based token interaction~\cite{vaswani2017attention}, and the Vision Transformer applies the same principle to image patches~\cite{dosovitskiy2021vit}. Pyramid Vision Transformer and PVTv2 adapt this representation to dense prediction through hierarchical, multi-resolution features~\cite{wang2021pvt,wang2022pvtv2}. COD models build on this hierarchy to reason globally while retaining progressively decoded detail~\cite{yang2021ugtr,huang2023fspnet,yin2022camoformer}. In parallel, self-supervised DINO reveals semantic organization in attention maps~\cite{caron2021dino}, and DINOv2 and DINOv3 scale transferable visual features while emphasizing dense representation quality~\cite{oquab2024dinov2,simeoni2025dinov3}. Segment Anything provides a promptable segmentation foundation model~\cite{kirillov2023sam}, and depth-enhanced adaptations illustrate how additional task cues can specialize such generic representations for COD~\cite{yu2024samdepth}. Broad semantic coverage, however, does not itself identify the category designated by a particular reference set. The relevant foundation tokens must be selected with target-conditioned evidence.

\section{Method}
\label{sec:method}

Given a query image $I_q\in\mathbb{R}^{C_q\times H_q\times W_q}$, reference-guided camouflaged object detection seeks to identify only the camouflaged objects specified by a set of visual references, rather than all plausible camouflaged foreground regions in the query~\cite{zhang2025refcod,wu2025uat}. Let $\mathcal{R}=\{I_r^k\}_{k=1}^{K}$ denote the reference set, where each $I_r^k\in\mathbb{R}^{C_r\times H_r\times W_r}$ contains a salient object belonging to the same target category $c$. The query image may contain one or more camouflaged instances of category $c$ amid visually similar background regions. The desired output is a binary mask $Y\in\{0,1\}^{H_q\times W_q}$, where $Y_{u,v}=1$ if pixel $(u,v)$ belongs to a target instance of category $c$ and $Y_{u,v}=0$ otherwise. We formulate the prediction as
\begin{equation}
P=f_{\theta}(I_q,\mathcal{R}),
\qquad P\in[0,1]^{H_q\times W_q},
\label{eq:problem-definition}
\end{equation}
where $f_{\theta}$ is the learnable Ref-COD model and $P$ is the predicted foreground-probability map. During training, $P$ is supervised by $Y$ together with the auxiliary and boundary targets defined later in this section. During inference, the references specify the semantic target and the model predicts its spatial extent in the query image.

\begin{figure}[!t]
    \centering
    \includegraphics[width=\textwidth]{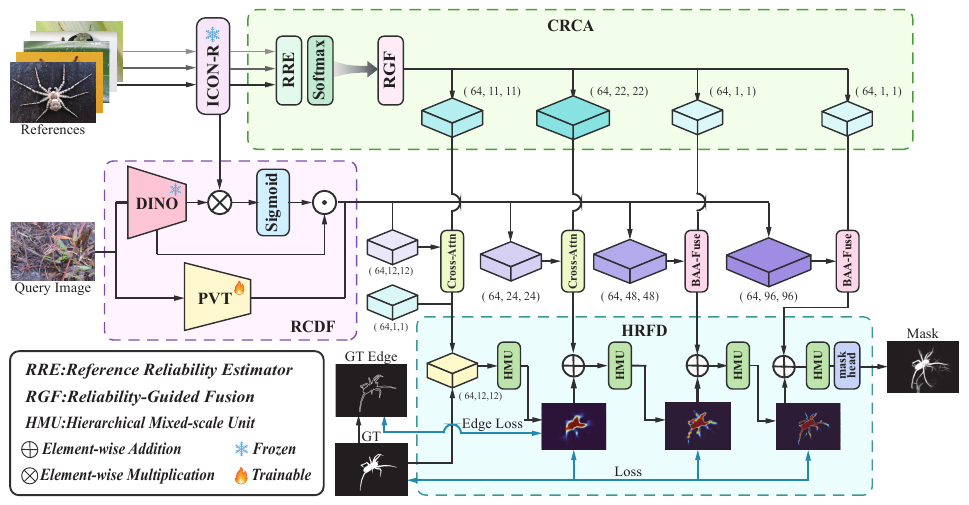}
    \caption{Overview of the consensus-aware multi-source fusion framework. CRCA aggregates heterogeneous references into consensus priors. RCDF couples the trainable PVTv2 and frozen DINOv3 backbones by using reference-conditioned correlation to gate foundation features before multi-scale fusion. The HRFD stage performs spatial cross-attention at deep stages, vector-based modulation at shallow stages, and HMU decoding to produce the final mask and auxiliary predictions.}
    \label{fig:pipeline}
\end{figure}

Figure~\ref{fig:pipeline} illustrates the overall architecture of the proposed framework. Our framework implements $f_{\theta}$ through consensus-aware multi-source fusion. A trainable PVTv2 encoder and a frozen DINOv3 encoder form the dual backbones of RCDF, producing hierarchical query features and patch-token features, respectively. ICON-R supplies a deep spatial feature, a middle spatial feature, and a global vector for each reference. RCDF averages the unprojected global reference vectors and uses the resulting target descriptor to gate DINOv3 features before multi-scale fusion with the PVTv2 features. CRCA independently aggregates the projected reference features, whose spatial and vector consensus priors subsequently condition the fused query features at matched semantic depths.

\subsection{Reference feature extraction}

We use ICON-R, a frozen salient-reference encoder based on the Integrity Cognition Network (ICON)~\cite{zhuge2023icon}. The original ICON decoder predicts an integrity-aware saliency map. Our reference wrapper additionally exposes deep and intermediate encoder features and uses the predicted soft foreground mask to suppress background responses. For reference image $I_r^k$, let $M^k=\sigma(S_{\mathrm{ICON}}(I_r^k))$ be the soft saliency mask, and let $E_d^k$ and $E_m^k$ denote the deep and intermediate encoder features. We compute
\begin{equation}
R_d^k=E_d^k\odot\mathcal{D}_d(M^k),
\qquad
R_m^k=E_m^k\odot\mathcal{D}_m(M^k),
\label{eq:icon-spatial}
\end{equation}
where $\mathcal{D}_j$ bilinearly resizes the mask to the corresponding feature resolution. A masked average of the deep feature provides the global reference vector:
\begin{equation}
r_v^k=
\frac{\sum_u \mathcal{D}_d(M^k)_u E_{d,u}^k}
     {\sum_u \mathcal{D}_d(M^k)_u+\epsilon}.
\label{eq:icon-vector}
\end{equation}
These foreground-aware representations retain localized structure and category-level semantics for consensus aggregation.

\subsection{Cross-reference consensus aggregation}

\begin{figure}[!t]
    \centering
    \includegraphics[width=\textwidth]{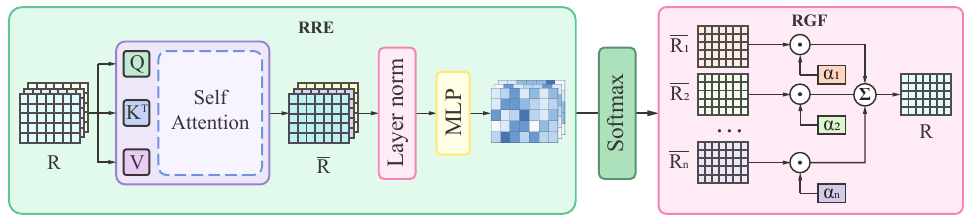}
    \caption{Internal structure of cross-reference consensus aggregation. The Reference Reliability Estimator (RRE) applies self-attention across the reference dimension and predicts reliability scores after normalization and projection. Reference-Guided Fusion (RGF) then uses the normalized weights $\{\alpha_k\}_{k=1}^{K}$ to aggregate the reference features $\{R_k\}_{k=1}^{K}$ into the consensus representation $\bar{R}$.}
    \label{fig:crca-detail}
\end{figure}

To address the third challenge, namely the heterogeneous quality and reliability of reference samples, we propose Cross-Reference Consensus Aggregation (CRCA) to construct robust target priors. Instead of assigning equal importance to references that may differ in pose, scale, appearance, or foreground clarity, CRCA explicitly models their interactions and learns how much each sample should contribute to the shared target representation. As illustrated in Figure~\ref{fig:crca-detail}, CRCA integrates a Reference Reliability Estimator (RRE) with Reference-Guided Fusion (RGF). RRE captures agreement and disagreement across references, while RGF uses the estimated reliability to preserve consistent target evidence and reduce the influence of less informative samples.

Given a projected reference tensor $X\in\mathbb{R}^{B\times K\times C\times H\times W}$, RRE treats the $K$ references at each spatial position as a sequence. It reshapes $X$ into $BHW$ sequences of length $K$ and applies multi-head self-attention followed by a residual projection. Let $\widetilde{X}_k$ denote the interaction-aware representation of reference $k$. A learned quality head produces the normalized reliability weights
\begin{equation}
\alpha_k = \operatorname{softmax}_k\left(g\left(\operatorname{LN}(\widetilde{X}_k)\right)\right),
\qquad
\bar{X}=\sum_{k=1}^{K}\alpha_k\widetilde{X}_k,
\label{eq:crca}
\end{equation}
where $g$ denotes the quality head. RGF then forms the weighted consensus $\bar{X}$ in Eq.~\ref{eq:crca}. We apply the proposed aggregation jointly to deep spatial features, middle spatial features, and projected reference vectors, producing the multi-granularity consensus priors $\bar{R}_d$, $\bar{R}_m$, and $\bar{R}_v$. By estimating reliability separately at these semantic levels, CRCA directly resolves the third challenge: it preserves localized structure and category-level semantics that are consistent across references while limiting the propagation of unreliable sample-specific cues.

\subsection{Reference-conditioned fusion}

The first challenge is target ambiguity in multi-object and distractor-rich scenes. To address it, Reference-Conditioned Dual-Backbone Fusion (RCDF) couples the trainable PVTv2 query backbone with the frozen DINOv3 foundation backbone. The two encoders provide complementary representations: PVTv2 learns task-specific hierarchical features that retain local structures, whereas DINOv3 supplies semantically discriminative patch representations learned from large-scale pretraining. Generic foundation features alone cannot distinguish target instances from semantically unrelated concealed objects, and direct fusion may therefore transfer strong but target-irrelevant responses. RCDF uses the references to select target-consistent DINOv3 evidence before it enters the multi-scale query hierarchy.

Let $r=K^{-1}\sum_{k=1}^{K}r_v^k$ be the mean unprojected foreground-aware reference vector and let $\{d_i\}_{i=1}^{N}$ be the DINOv3 patch tokens. Averaging the foreground-aware vectors summarizes the target semantics shared by the references, while retaining the original ICON-R representation space for correlation estimation. RCDF maps the reference descriptor and patch tokens to a common embedding space, applies $\ell_2$ normalization, and computes a sigmoid correlation score for each patch:
\begin{equation}
c_i=\sigma\left(\tau\frac{\phi_r(r)^\top\phi_d(d_i)}{\lVert\phi_r(r)\rVert_2\lVert\phi_d(d_i)\rVert_2}\right),
\label{eq:corr}
\end{equation}
where $\phi_r$ and $\phi_d$ are learned linear projections and $\tau$ is a learnable scale controlling the sharpness of target selection. The scores $\{c_i\}_{i=1}^{N}$ are rearranged according to the DINOv3 patch grid to form a spatial correlation map $C$. A high response indicates that the local query representation is semantically consistent with the reference target, whereas a low response attenuates foundation features associated with background regions or unrelated objects.

For the $j$-th PVTv2 stage, let $Q_j$ denote the projected query feature and let $D$ denote the DINOv3 spatial feature reconstructed from its patch tokens. We resize $D$ and $C$ to the spatial resolution of $Q_j$, project the resized foundation feature to the common model dimension, and perform gated convolutional fusion as
\begin{equation}
F_j=\psi_j\left(\left[Q_j,\;\pi_j\left(\mathcal{I}_j(D)\right)\odot\mathcal{I}_j(C)\right]\right), \qquad j\in\{0,1,2,3\},
\label{eq:rcdf-fusion}
\end{equation}
where $\mathcal{I}_j$ denotes bilinear interpolation to the $j$-th feature resolution, $\pi_j$ is the corresponding DINOv3 feature projection, $[\cdot,\cdot]$ denotes channel-wise concatenation, $\odot$ denotes element-wise multiplication, and $\psi_j$ is a stage-specific convolutional fusion block. Applying the reference-conditioned map at every scale allows deep features to retain target-level semantics while shallow features preserve spatial details needed for subsequent decoding.

RCDF performs patch-wise selection rather than collapsing the query into a single target location. Consequently, spatially disconnected regions can receive high correlation scores whenever they share the semantics specified by the references. This property is particularly important in multi-object scenes: multiple camouflaged instances can be retained simultaneously, while visually salient but reference-inconsistent regions are suppressed before hierarchical reference interaction and decoding. The frozen DINOv3 branch provides stable semantic evidence, and the trainable PVTv2 and fusion blocks adapt this evidence to the localization requirements of Ref-COD. RCDF thus resolves the first challenge by converting a generic foundation representation into target-conditioned, spatially distributed evidence suitable for both multi-instance recovery and distractor suppression.

\subsection{Hierarchical reference fusion and decoding}

To address the second challenge of recovering small regions, thin structures, and ambiguous boundaries, we design hierarchical reference fusion and decoding (HRFD) to inject each consensus prior at a query depth that matches its semantic granularity. Uniform reference injection can impose coarse spatial correspondence on high-resolution features or provide insufficient localization cues to deep features, which weakens the reconstruction of fine target details. HRFD therefore uses spatial reference priors for deep semantic interaction and the global vector prior for shallow feature modulation, allowing target structure and category semantics to complement each other throughout decoding.

At the deeper levels, the fused query features attend to $\bar{R}_d$ and $\bar{R}_m$, respectively. Our cross-attention block projects query, key, and value tokens, applies a scaled residual connection to the attended output, and follows it with a feed-forward residual block. At the shallow levels, HRFD uses broadcast attention adaptation (BAA) with $\bar{R}_v$ to modulate high-resolution features without forcing a coarse spatial correspondence.

To maintain target awareness during progressive decoding, HRFD generates a relevance mask from the deepest feature and $\bar{R}_v$ and appends it to every decoding stage. Hierarchical mixed-scale units (HMUs) decode the conditioned features in a top-down manner, while ASPP enriches the deepest query projection. This coordinated design produces a final prediction $P$, auxiliary predictions $\{P_j\}_{j=1}^{J}$, and an edge prediction $E$. The training objective is
\begin{equation}
\mathcal{L}=\mathcal{L}_{\mathrm{str}}(P,Y)+\lambda_1\sum_{j=1}^{J}\mathcal{L}_{\mathrm{str}}(P_j,Y)+\lambda_2\mathcal{L}_{\mathrm{edge}}(E,Y_{\mathrm{edge}}),
\label{eq:loss}
\end{equation}
The depth-matched interactions preserve target semantics during feature refinement, while progressive decoding and boundary supervision recover local structures at increasing spatial resolutions. HRFD therefore resolves the second challenge by coordinating semantic guidance, multi-scale reconstruction, and explicit boundary learning instead of treating detail recovery as an isolated final-stage operation.

\section{Experiments}
\label{sec:experiments}

\subsection{Dataset and evaluation metrics}

We conduct all experiments on R2C7K~\cite{zhang2025refcod}, a large-scale benchmark for reference-guided camouflaged object detection. R2C7K contains 6,615 images spanning 64 object categories, including 1,600 salient reference images and 5,015 camouflaged query images. Each category provides 25 reference images collected from real-world scenes. Following the official protocol, 20 reference images per category are assigned to training and the remaining five to testing. The camouflaged-image training set includes the training samples from COD10K, while evaluation uses the corresponding test samples from CAMO and NC4K. The test set is reported as Overall and is further divided into Single-object and Multi-object subsets according to the number of camouflaged objects present in each scene.

We evaluate predictions using S-measure ($S_m$)~\cite{fan2017structure}, adaptive E-measure ($\alpha E$)~\cite{fan2018enhanced}, weighted F-measure ($wF$)~\cite{margolin2014evaluate}, and mean absolute error (MAE). Higher values indicate better performance for $S_m$, $\alpha E$, and $wF$, whereas a lower MAE is preferred.

\subsection{Implementation details}

Our implementation uses PVTv2-B2 as the trainable query encoder, DINOv3 ViT-B/16 as the frozen foundation encoder, and the ResNet-50 variant of ICON-R as the frozen reference encoder. Reference images are resized to $352\times352$ for ICON-R, whereas query images are resized to $384\times384$ for PVTv2 and DINOv3 during both training and testing. The model dimension is 64, and optimization uses SGD with momentum 0.9 and weight decay $5\times10^{-4}$. Training uses 40 epochs, a batch size of 8, an initial learning rate of 0.05 for non-backbone parameters and 0.005 for backbone parameters, three warm-up epochs, and linear decay. The selected configuration uses two deep and two shallow interaction levels, four HMUs, three auxiliary predictions, an auxiliary-loss weight of 0.5, an edge-loss weight of 0.2, a cross-attention residual scale of 0.3, four cross-attention heads, a correlation-scale initialization of 20, eight HMU groups with a hidden dimension of 48, a mask-guidance coefficient of 0.25, and ASPP dilation rates of $(3,6,9)$. We use a 5-shot setting during both training and testing: five reference samples are randomly selected for each query during training, and five references are used for each query at inference. The frozen ICON-R encoder processes all reference images offline before model training; its deep, middle, and vector features are stored as \texttt{.npz} files and reused unchanged during training and testing. No test-time augmentation (TTA) is used for the reported comparison, ablation, or parameter-sensitivity results.

\subsection{Quantitative comparison}

Table~\ref{tab:comparison-overall} compares five methods on the overall, single-object, and multi-object subsets. On the overall set, our complete model achieves the best results across all four metrics, with 0.895 $S_m$, 0.941 $\alpha E$, 0.825 $wF$, and 0.017 MAE. Compared with the strongest competing method, RefOnce~\cite{wu2025refonce}, these results improve $S_m$, $\alpha E$, and $wF$ by 0.005, 0.004, and 0.006, respectively, while reducing MAE by 0.002. The consistent changes across structural similarity, alignment, weighted foreground quality, and pixel error indicate that the improvement is not confined to a single evaluation criterion.

The subset results further reveal the behavior under different scene complexities. On the single-object subset, our method obtains 0.899 $S_m$, 0.943 $\alpha E$, 0.830 $wF$, and 0.016 MAE. The multi-object subset is more challenging for all compared methods, yet our method retains the strongest $S_m$, $wF$, and MAE and ties RefOnce for the best $\alpha E$. Relative to RefOnce on this subset, it improves $S_m$ by 0.006 and $wF$ by 0.013 while reducing MAE by 0.003. This larger gain in $wF$ is consistent with the intended role of RCDF: reference-conditioned patch selection retains spatially separated target instances while suppressing reference-inconsistent regions. CRCA and HRFD complement this selection by stabilizing the reference cues and refining target structures during decoding.

Among methods using PVTv2-B2 as the query backbone, our model improves upon UAT by 0.040 $S_m$, 0.029 $\alpha E$, and 0.068 $wF$ on the overall set, together with a 0.009 reduction in MAE. On the multi-object subset, the corresponding improvements are 0.054, 0.030, and 0.079, with MAE reduced by 0.012. These comparisons suggest that the gains arise from how reference evidence is aggregated, used to condition foundation features, and injected across semantic depths rather than from the query backbone alone. The Qwen+SAM baseline combines Qwen3-VL with SAM2, but its general-purpose semantic and segmentation capabilities do not provide the same target-specific consistency in these camouflaged scenes.

\begin{table}[!htbp]
\caption{Quantitative comparison on the overall, single-object, and multi-object subsets. Higher values are better for $S_m$, $\alpha E$, and $wF$; lower values are better for MAE.}
\label{tab:comparison-overall}
\centering
\scriptsize
\setlength{\tabcolsep}{3pt}
\resizebox{\textwidth}{!}{%
\begin{tabular}{ll*{12}{c}}
\toprule
Method & Backbone
& \multicolumn{4}{c}{Overall}
& \multicolumn{4}{c}{Single-object}
& \multicolumn{4}{c}{Multi-object} \\
\cmidrule(lr){3-6}\cmidrule(lr){7-10}\cmidrule(lr){11-14}
& & $S_m\uparrow$ & $\alpha E\uparrow$ & $wF\uparrow$ & MAE$\downarrow$
& $S_m\uparrow$ & $\alpha E\uparrow$ & $wF\uparrow$ & MAE$\downarrow$
& $S_m\uparrow$ & $\alpha E\uparrow$ & $wF\uparrow$ & MAE$\downarrow$ \\
\midrule
R2CNet & ResNet-50
& 0.805 & 0.879 & 0.669 & 0.036
& 0.810 & 0.880 & 0.674 & 0.035
& 0.747 & 0.872 & 0.602 & 0.046 \\
UAT & PVTv2-B2
& 0.855 & 0.912 & 0.757 & 0.026
& 0.859 & 0.913 & 0.761 & 0.025
& 0.805 & 0.900 & 0.701 & 0.033 \\
Qwen+SAM & --
& 0.852 & 0.917 & 0.793 & 0.029
& 0.861 & 0.926 & 0.805 & 0.027
& 0.778 & 0.844 & 0.683 & 0.047 \\
RefOnce & PVTv2-B2
& 0.890 & 0.937 & 0.819 & 0.019
& 0.894 & 0.937 & 0.825 & 0.018
& 0.853 & \textbf{0.930} & 0.767 & 0.024 \\
Ours / All components & PVTv2-B2
& \textbf{0.895} & \textbf{0.941} & \textbf{0.825} & \textbf{0.017}
& \textbf{0.899} & \textbf{0.943} & \textbf{0.830} & \textbf{0.016}
& \textbf{0.859} & \textbf{0.930} & \textbf{0.780} & \textbf{0.021} \\
\bottomrule
\end{tabular}
}
\end{table}

\subsection{Qualitative comparison}

\begin{figure}[!t]
    \centering
    \includegraphics[width=\textwidth]{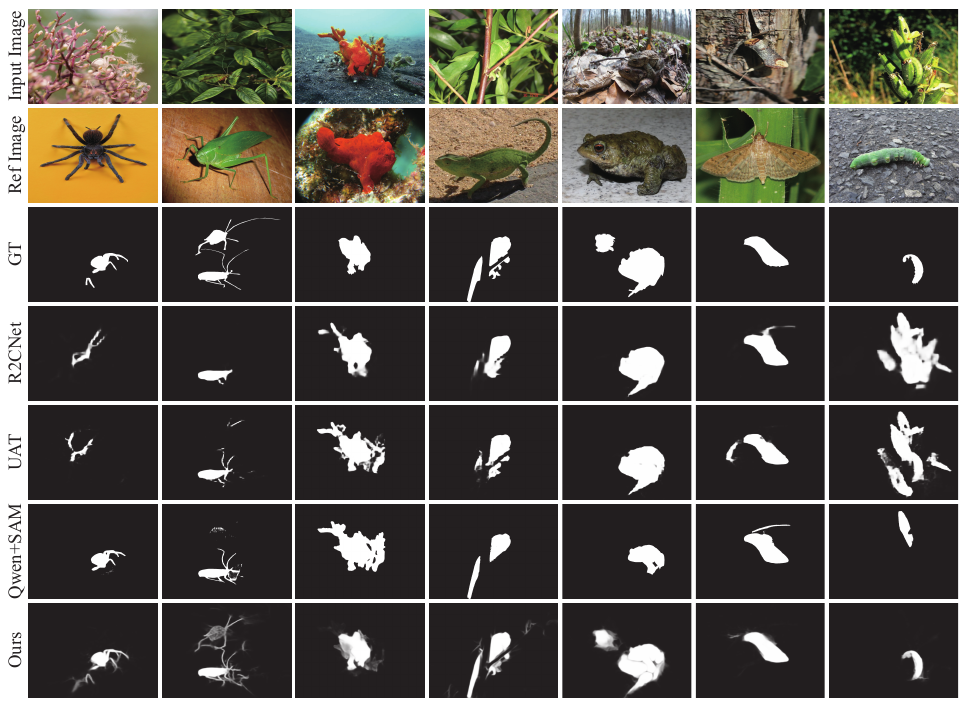}
    \caption{Qualitative comparison on representative query--reference pairs. From top to bottom, the rows show the query image, reference image, ground truth, and predictions from R2CNet, UAT, Qwen+SAM, and our method.}
    \label{fig:qualitative}
\end{figure}

Figure~\ref{fig:qualitative} compares R2CNet, UAT, Qwen+SAM, and our method on representative query--reference pairs. The examples cover low-contrast targets, textured backgrounds, and thin structures. Across these cases, our method produces more coherent target masks while suppressing background responses that remain in the comparison outputs. This qualitative evidence complements the quantitative results in Table~\ref{tab:comparison-overall}.

\subsection{Factorial component ablation}

Table~\ref{tab:component-ablation} reports a full factorial ablation over CRCA, RCDF, and hierarchical reference fusion and decoding (HRFD). Each component improves the baseline when introduced independently. CRCA increases the overall $S_m$ and $wF$ from 0.8550 and 0.7570 to 0.8610 and 0.7660, respectively; RCDF produces the largest individual gains, reaching 0.8720 $S_m$ and 0.7850 $wF$; and HRFD reaches 0.8660 $S_m$ and 0.7740 $wF$. Pairwise combinations improve further, with RCDF+HRFD giving the strongest two-component result across all four overall metrics. The complete model achieves 0.8950 $S_m$, 0.9410 $\alpha E$, 0.8250 $wF$, and 0.0170 MAE, improving the baseline by 0.0400, 0.0290, and 0.0680 on the three higher-is-better metrics while reducing MAE by 0.0090. The consistent improvements on both the single-object and multi-object subsets indicate that the three components provide complementary evidence rather than redundant refinements.

\begin{table}[!b]
\caption{Full-factorial ablation of CRCA, RCDF, and HRFD on the Overall, Single-object, and Multi-object subsets. Higher values are better for $S_m$, $\alpha E$, and $wF$; lower values are better for MAE.}
\label{tab:component-ablation}
\centering
\scriptsize
\setlength{\tabcolsep}{3pt}
\resizebox{\textwidth}{!}{%
\begin{tabular}{ccc*{12}{c}}
\toprule
CRCA & RCDF & HRFD
& \multicolumn{4}{c}{Overall}
& \multicolumn{4}{c}{Single-object}
& \multicolumn{4}{c}{Multi-object} \\
\cmidrule(lr){4-7}\cmidrule(lr){8-11}\cmidrule(lr){12-15}
& & & $S_m\uparrow$ & $\alpha E\uparrow$ & $wF\uparrow$ & MAE$\downarrow$
& $S_m\uparrow$ & $\alpha E\uparrow$ & $wF\uparrow$ & MAE$\downarrow$
& $S_m\uparrow$ & $\alpha E\uparrow$ & $wF\uparrow$ & MAE$\downarrow$ \\
\midrule
$-$ & $-$ & $-$
& 0.8550 & 0.9120 & 0.7570 & 0.0260
& 0.8590 & 0.9130 & 0.7610 & 0.0250
& 0.8050 & 0.9000 & 0.7010 & 0.0330 \\
$\checkmark$ & $-$ & $-$
& 0.8610 & 0.9160 & 0.7660 & 0.0247
& 0.8654 & 0.9172 & 0.7722 & 0.0239
& 0.8210 & 0.9050 & 0.7100 & 0.0315 \\
$-$ & $\checkmark$ & $-$
& 0.8720 & 0.9230 & 0.7850 & 0.0229
& 0.8762 & 0.9242 & 0.7905 & 0.0222
& 0.8340 & 0.9120 & 0.7350 & 0.0288 \\
$-$ & $-$ & $\checkmark$
& 0.8660 & 0.9190 & 0.7740 & 0.0238
& 0.8703 & 0.9202 & 0.7800 & 0.0231
& 0.8270 & 0.9080 & 0.7200 & 0.0300 \\
$\checkmark$ & $\checkmark$ & $-$
& 0.8810 & 0.9320 & 0.8060 & 0.0208
& 0.8847 & 0.9331 & 0.8111 & 0.0202
& 0.8480 & 0.9220 & 0.7600 & 0.0258 \\
$\checkmark$ & $-$ & $\checkmark$
& 0.8760 & 0.9270 & 0.7950 & 0.0219
& 0.8799 & 0.9281 & 0.8002 & 0.0213
& 0.8410 & 0.9170 & 0.7480 & 0.0273 \\
$-$ & $\checkmark$ & $\checkmark$
& 0.8870 & 0.9360 & 0.8160 & 0.0189
& 0.8903 & 0.9371 & 0.8210 & 0.0184
& 0.8570 & 0.9260 & 0.7710 & 0.0233 \\
$\checkmark$ & $\checkmark$ & $\checkmark$
& \textbf{0.8950} & \textbf{0.9410} & \textbf{0.8250} & \textbf{0.0170}
& \textbf{0.8990} & \textbf{0.9422} & \textbf{0.8300} & \textbf{0.0166}
& \textbf{0.8590} & \textbf{0.9300} & \textbf{0.7800} & \textbf{0.0208} \\
\bottomrule
\end{tabular}
}
\end{table}

\begin{figure}[!t]
    \centering
    \includegraphics[width=\linewidth]{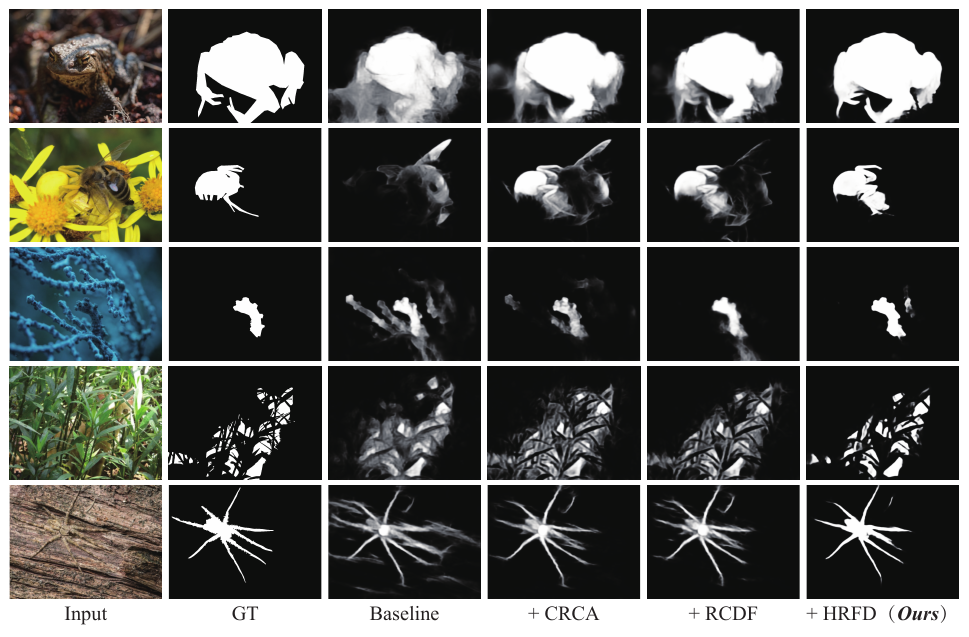}
    \caption{Qualitative visualization of progressive component variants. The labels embedded in the figure denote the progressive variants and are distinct from the factorial combinations evaluated in Table~\ref{tab:component-ablation}.}
    \label{fig:ablation}
\end{figure}

\subsection{Ablation study}

\begin{figure}[!t]
    \centering
    \includegraphics[width=\textwidth]{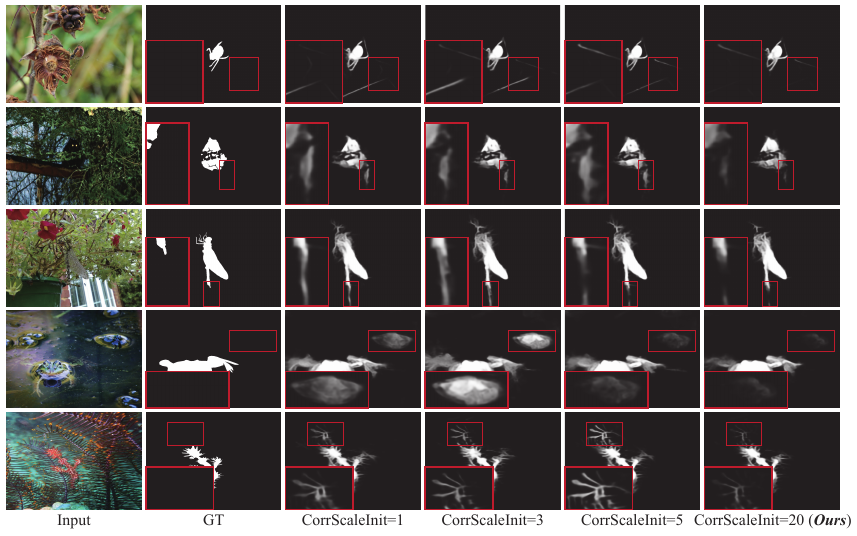}
    \caption{Qualitative comparison of correlation-scale initialization. Columns show the input, ground truth, scales of 1, 3, and 5, and the selected scale of 20 (Ours). Red boxes enlarge challenging regions and boundaries.}
    \label{fig:corr-scale-visual}
\end{figure}

\begin{figure}[!t]
    \centering
    \includegraphics[width=\textwidth]{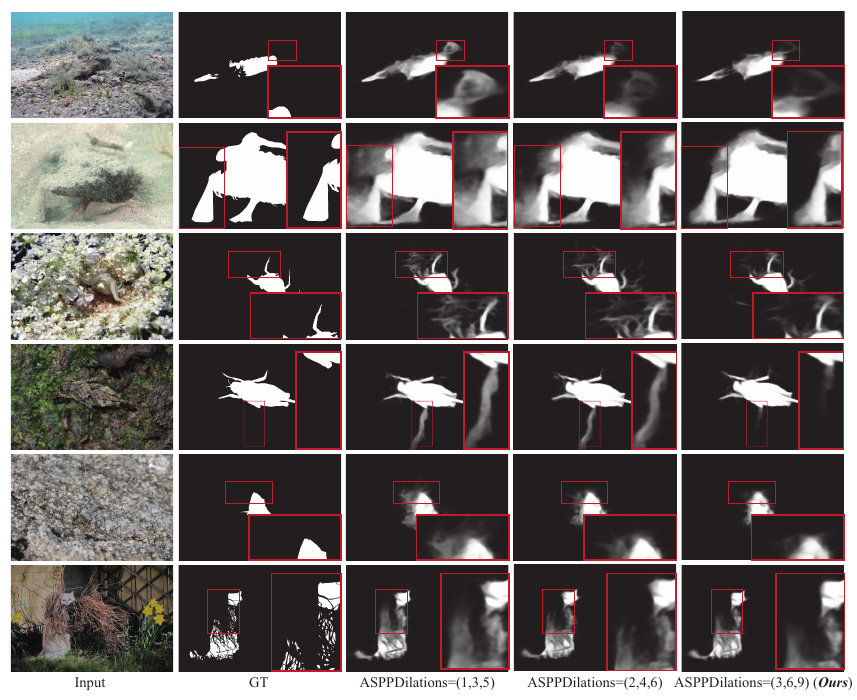}
    \caption{Qualitative comparison of ASPP dilation settings. Columns show the input, ground truth, groups $(1,3,5)$ and $(2,4,6)$, and the selected group $(3,6,9)$ (Ours). Red boxes enlarge challenging regions and boundaries.}
    \label{fig:aspp-visual}
\end{figure}

Table~\ref{tab:parameter} reports three representative one-factor parameter studies covering the supervision balance, foundation-feature gating, and multi-scale context configuration. Each study starts from the same complete-model configuration and changes only the parameter named in that block while keeping all other settings fixed. Consequently, the selected row in every block represents the same complete-model run and intentionally repeats the same performance values. Figures~\ref{fig:corr-scale-visual} and~\ref{fig:aspp-visual} provide qualitative comparisons for the correlation-scale initialization and ASPP dilation settings, respectively. These studies are separate from the factorial design in Table~\ref{tab:component-ablation} and serve as controlled configuration-selection evidence rather than as a joint component ablation.

The parameter trends also clarify how the three design choices affect different aspects of prediction. For edge supervision, a weight of 0.20 provides the most balanced result: reducing the weight weakens boundary guidance, whereas an excessively large weight overemphasizes local contours at the expense of region-level consistency. The correlation-scale study shows that small initial scales produce less selective query--reference responses. As illustrated in Fig.~\ref{fig:corr-scale-visual}, these settings often yield diffuse activations around thin structures or incomplete responses over the target extent, while the selected scale produces more concentrated foreground evidence and cleaner boundaries. The dilation study exhibits a complementary pattern. Closely spaced small rates offer limited context variation, whereas the selected $(3,6,9)$ group combines local detail with a broader receptive field. Fig.~\ref{fig:aspp-visual} consequently shows fewer fragmented regions and more complete recovery of low-contrast targets. These observations support using the selected settings together: edge supervision refines contours, correlation scaling sharpens reference-conditioned selection, and the dilation group supplies the multi-scale context needed to preserve object extent.

\begin{table}[!htbp]
\caption{Three representative one-factor parameter studies. Each block changes only the named setting, while selected rows repeat the shared complete-model result. Higher values are better for $S_m$, $\alpha E$, and $wF$; lower values are better for MAE.}
\label{tab:parameter}
\centering
\scriptsize
\setlength{\tabcolsep}{4pt}
\begin{tabular}{llcccc}
\toprule
Study & Setting & $S_m\uparrow$ & $\alpha E\uparrow$ & $wF\uparrow$ & MAE$\downarrow$ \\
\midrule
Edge-loss weight & 0.05 & 0.8884 & 0.9402 & 0.8162 & 0.0174 \\
& 0.10 & 0.8904 & 0.9334 & 0.8108 & 0.0177 \\
& 0.50 & 0.8885 & 0.9318 & 0.8192 & 0.0170 \\
& \textbf{0.20 (selected)} & \textbf{0.8946} & \textbf{0.9414} & \textbf{0.8253} & \textbf{0.0168} \\
\midrule
Correlation-scale initialization & 1 & 0.8874 & 0.9368 & 0.8172 & 0.0172 \\
& 3 & 0.8873 & 0.9394 & 0.8170 & 0.0172 \\
& 5 & 0.8876 & 0.9366 & 0.8145 & 0.0176 \\
& \textbf{20 (selected)} & \textbf{0.8946} & \textbf{0.9414} & \textbf{0.8253} & \textbf{0.0168} \\
\midrule
ASPP dilation rates & $(1,3,5)$ & 0.8875 & 0.9386 & 0.8156 & 0.0182 \\
& $(2,4,6)$ & 0.8890 & 0.9329 & 0.8112 & 0.0179 \\
& \textbf{$(3,6,9)$ (selected)} & \textbf{0.8946} & \textbf{0.9414} & \textbf{0.8253} & \textbf{0.0168} \\
\bottomrule
\end{tabular}
\end{table}

\subsection{Computational efficiency}

We measure computational efficiency on one NVIDIA GeForce RTX 4090 D GPU with 24~GB of memory. Table~\ref{tab:efficiency} summarizes the resulting computational profile. With a batch size of 8, one training iteration takes 140.68~ms on average. Under FP32 inference with a $384\times384$ query and five references, the model processes one query in 24.31~ms, corresponding to 41.13 frames per second (FPS). This inference measurement uses the offline-cached ICON-R reference features described above and therefore excludes reference-feature extraction. The model requires 143.54~GFLOPs (71.77~GMACs) and contains 115.67 million parameters, of which 30.00 million are trainable and 85.67 million remain frozen throughout optimization.

This profile reflects the intended separation between reusable reference processing and query-time prediction. Reference features can be computed once and reused for multiple queries, so the online path focuses on reference aggregation, query--reference interaction, and progressive decoding. The resulting throughput indicates that the additional reference-guided reasoning remains practical for batch evaluation and interactive inspection. Moreover, freezing most parameters confines optimization to the task-specific components, reducing the trainable portion of the model while retaining the representation capacity of the pretrained encoders. The reported latency should therefore be interpreted as the cost of processing a query after reference preparation; applications that replace or frequently update the reference set must additionally account for the one-time reference-feature extraction cost.

\begin{table}[!htbp]
\caption{Computational efficiency on one NVIDIA GeForce RTX 4090 D. Inference uses FP32, a $384\times384$ query, five references, and cached ICON-R features; latency excludes reference-feature extraction.}
\label{tab:efficiency}
\centering
\small
\setlength{\tabcolsep}{6pt}
\begin{tabular}{ll}
\toprule
Item & Measurement \\
\midrule
Training iteration time (batch size 8) & 140.68 ms/step \\
Inference latency (FP32, 5-shot) & 24.31 ms/image \\
Inference throughput & 41.13 FPS \\
Computational cost & 143.54 GFLOPs / 71.77 GMACs \\
Peak inference memory & 548.6 MiB \\
Peak training memory & 7008.8 MiB \\
\bottomrule
\end{tabular}
\end{table}

\section{Conclusion and Limitation}
\label{sec:conclusion}

We presented a consensus-aware multi-source fusion framework for reference-guided camouflaged object detection. CRCA aggregates heterogeneous references into reliability-weighted spatial and vector priors, RCDF uses reference-conditioned correlation to select target-relevant DINOv3 features, and HRFD injects reference evidence at matched semantic depths before progressive decoding. The full-factorial ablation verifies that each component improves the baseline independently and that their combinations yield cumulative gains, with RCDF providing the strongest individual contribution and the complete model performing best across the overall metrics. Parameter studies further identify a consistent operating configuration, while qualitative comparisons show improved localization and boundary recovery on challenging query--reference pairs. Together, these results support reliability-aware reference aggregation, target-conditioned foundation-feature selection, and depth-matched decoding as complementary mechanisms for reference-guided camouflaged object detection. A limitation is that the evaluation uses a single dataset; future work will assess cross-dataset generalization under broader reference conditions.

\FloatBarrier
\section*{Acknowledgments}
This work was supported by the National Natural Science Foundation of China (Grant No. 62402152) and the Zhejiang Province Natural Science Foundation of China (Grant No. LQN25F020017).
\bibliographystyle{plainnat}
\bibliography{references}
\clearpage
\appendix
\setcounter{figure}{0}
\renewcommand{\thefigure}{A\arabic{figure}}
\section{Additional qualitative analyses}

The following visualizations provide mechanism-oriented qualitative evidence for the reference-conditioned feature selection and reference-guided refinement stages. Each panel uses representative examples to illustrate intermediate responses. They complement, rather than replace, the controlled factorial ablation reported in the main manuscript.

\subsection{Reference-conditioned DINOv3 feature selection}

\begin{figure}[!htbp]
    \centering
    \includegraphics[width=\textwidth,height=.65\textheight,keepaspectratio]{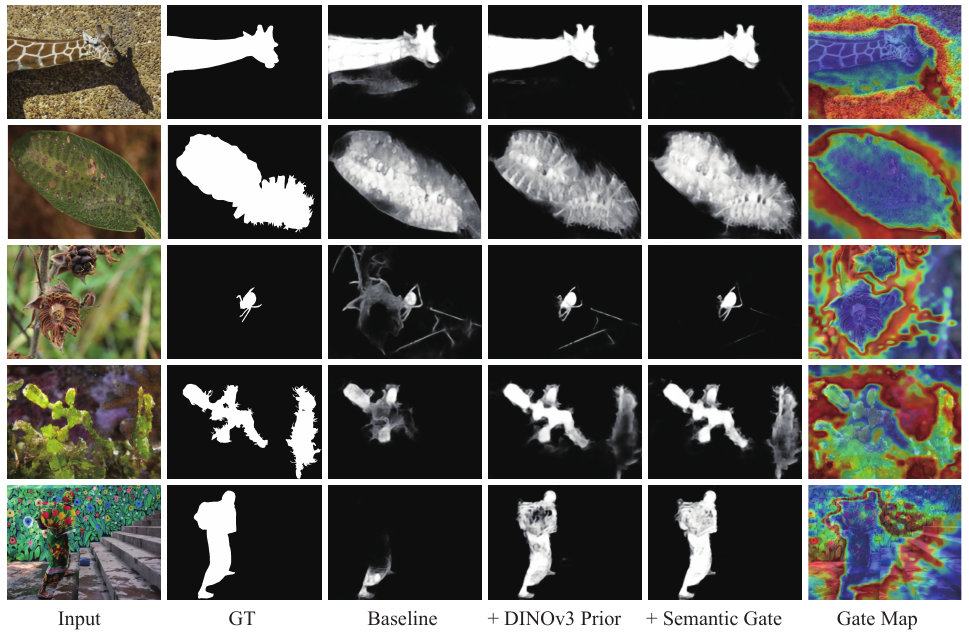}
    \caption{Qualitative visualization of reference-conditioned DINOv3 feature selection. From left to right, the columns show the input, ground truth (GT), baseline prediction, output with the DINOv3 prior, output after the Semantic Gate, and gate map. In the present formulation, the Semantic Gate denotes RCDF's reference-conditioned correlation map.}
    \label{fig:supp-semantic-gate}
\end{figure}

Figure~\ref{fig:supp-semantic-gate} illustrates how Reference-Conditioned Dual-Backbone Fusion (RCDF) transforms generic DINOv3 responses into target-conditioned evidence. The ungated DINOv3 prior can contain broad activations around semantically salient but target-ambiguous regions. The Semantic Gate, which corresponds to RCDF's reference-conditioned correlation map, attenuates responses that are inconsistent with the reference target while retaining target-aligned structures. The gate maps make this spatial selectivity explicit, with high responses concentrated around regions that agree with the reference-guided target. These examples are qualitative diagnostics and should be interpreted together with the controlled component ablation in the main manuscript.

\clearpage
\subsection{Progressive reference-guided refinement}

\begin{figure}[H]
    \centering
    \includegraphics[width=.95\textwidth,height=.65\textheight,keepaspectratio]{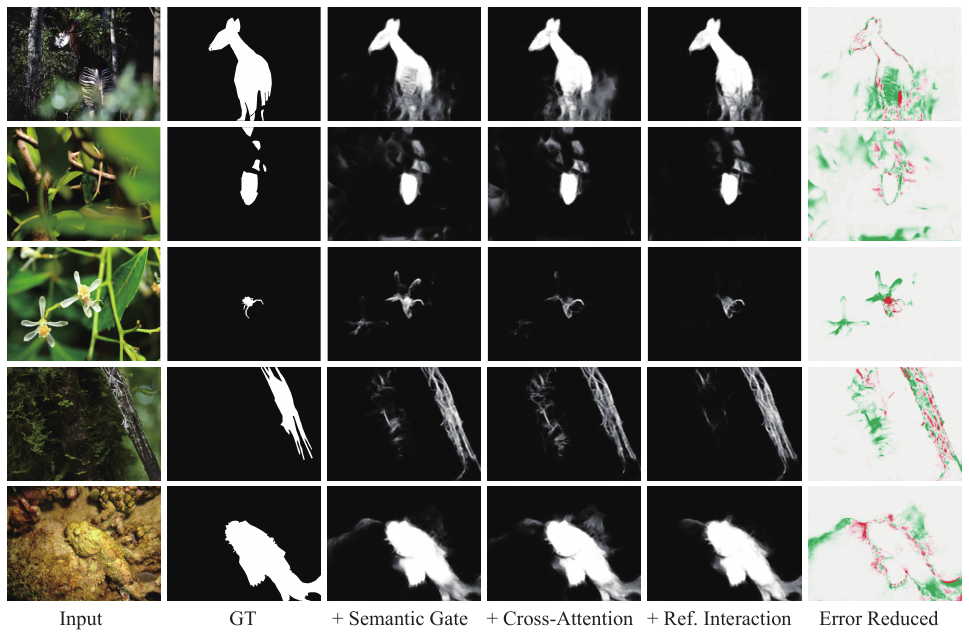}
    \caption{Qualitative visualization of progressive reference-guided refinement. From left to right, the columns show the input, GT, outputs after the Semantic Gate, Cross-Attention, and Ref. Interaction, followed by residual error. The latter two stages are the reference-guided interactions in HRFD.}
    \label{fig:supp-reference-refinement}
\end{figure}

Figure~\ref{fig:supp-reference-refinement} visualizes progressive refinement from the semantic-gated response to the final prediction. The Semantic Gate implements RCDF's correlation-based selection, whereas Cross-Attention and Ref. Interaction denote the reference-guided interactions in hierarchical reference fusion and decoding (HRFD). Later stages suppress extraneous activations, and the rightmost column shows residual errors relative to GT. This visualization qualitatively supports the complementary roles of target-conditioned selection and reference-guided refinement without introducing an independent ablation factor.

\end{document}